\documentclass[a4paper,UKenglish,cleveref, autoref, thm-restate]{lipics-v2021}

\hideLIPIcs

\nolinenumbers
\graphicspath{{./images/}}
\usepackage{graphicx, float, xcolor, tikz}
\definecolor{mygreen}{rgb}{0,0.6,0}

\lstdefinelanguage{eprime}
{
    keywords={forAll, forall, letting, union, max, sum, exists, int, of, find, given, flatten, matrix, indexed, by, be, domain, ->, /},
    keywordstyle=\color{blue}\bfseries,
    ndkeywords={atleast, atmost, gcc, allDiff},
    ndkeywordstyle=\color{purple}\bfseries,
    identifierstyle=\color{black},
    sensitive=false,
    comment=[l]{\$},
    commentstyle=\color{mygreen}\ttfamily,
    stringstyle=\color{red}\ttfamily,
    morestring=[b]',
    morestring=[b]"
}

\title{Towards Automatic Design of Factorio Blueprints}

\author{Sean Patterson}{School of Computer Science, University of St Andrews, UK}{seanapatterson@btinternet.com}{}{}

\author{Joan Espasa}{School of Computer Science, University of St Andrews, UK}{jea20@st-andrews.ac.uk}{https://orcid.org/0000-0002-9021-3047}{}

\author{Mun See Chang}{School of Computer Science, University of St Andrews, UK}{msc2@st-andrews.ac.uk}{https://orcid.org/0000-0003-2428-6130}{}

\author{Ruth Hoffmann}{School of Computer Science, University of St Andrews, UK}{rh347@st-andrews.ac.uk}{https://orcid.org/0000-0002-1011-5894}{}

\authorrunning{S. Patterson, J. Espasa, M.S. Chang and R. Hoffmann} 

\Copyright{Sean Patterson, Joan Espasa, Mun See Chang and Ruth Hoffmann} 



\EventEditors{John Q. Open and Joan R. Access}
\EventNoEds{2}
\EventLongTitle{42nd Conference on Very Important Topics (CVIT 2016)}
\EventShortTitle{CVIT 2016}
\EventAcronym{CVIT}
\EventYear{2016}
\EventDate{December 24--27, 2016}
\EventLocation{Little Whinging, United Kingdom}
\EventLogo{}
\SeriesVolume{42}
\ArticleNo{23}

\begin{document}

\maketitle

\begin{abstract}
Factorio is a 2D construction and management simulation video game about building automated factories to produce items of increasing complexity.
A core feature of the game is its blueprint system, which allows players to easily save and replicate parts of their designs. 
Blueprints can reproduce any layout of objects in the game, but are typically used to encapsulate a complex behaviour, such as the production of a non-basic object. Once created, these blueprints are then used as basic building blocks, allowing the player to create a layer of abstraction. The usage of blueprints not only eases the expansion of the factory but also allows the sharing of designs with the game's community.
The layout in a blueprint can be optimised using various criteria, such as the total space used or the final production throughput. The design of an optimal blueprint is a hard combinatorial problem, interleaving elements of many well-studied problems such as bin-packing, routing or network design.
This work presents a new challenging problem and explores the feasibility of a constraint model to optimise Factorio blueprints, balancing correctness, optimality, and performance.
\end{abstract}

\section{Introduction}
\label{sec:intro}
Factorio, produced by Wube Software~\cite{factorio}, is a game about building automated factories to produce items of increasing complexity, within an infinite 2D world. 
The player must optimise and expand their designs as the game progresses and their factory grows in both scope and scale.
The challenges players must solve to achieve this emulate many real-world problems, such as bin-packing, routing, network design and scheduling.
In the industry, these problems seldomly appear in isolation. For example, the core problem for a logistics company is the vehicle routing problem, which will get affected by other problems such as the scheduling of its driver workforce or by how many vehicles can be brought to the field.
Factorio provides a contrived testbed for modelling combinations of such problems.

Various elements of Factorio have previously been studied.
Leue~\cite{Leue} used Petri Nets and Linear Temporal Logic to verify the performance of `belt balancing' systems, allowing item flow in the factory to be equally distributed across multiple conveyor paths.
Reid et al.~\cite{Reid} compared several techniques 
to optimise the placement of conveyor belts from one point to another while avoiding obstacles, achieving results in an area of up to 12 by 12 tiles.
Their model aims to connect a single input and output, which is generally not representative of a complete factory.
While their findings are optimal for a single path, this optimality may not hold when multiple routes are competing for space. Finally, Duhan et al.~\cite{Duhan} showed interest in optimising factories using deep reinforcement learning.

One of the main game features is a \emph{blueprint} system, allowing players to save their factory designs.
Blueprints copy the layout of player-built objects in an area of up to 10,000 by 10,000 tiles.
Although this feature is useful in many in-game situations, its main use case is to replicate efficient designs for the production of a given non-basic game object. This allows complex parts of the factory to be abstracted, removing the need for the player to focus on irrelevant details.
Blueprint designs can also be uploaded to community websites and workshops for other players to download and use.


This paper explores the feasibility of automatically generating and optimising Factorio blueprints.
More concretely, we contribute a set of models in \textsc{Essence Prime}~\cite{essence-prime-description}, a well-known declarative constraint modelling language. Moreover, we take advantage of \textsc{Savile Row}~\cite{savilerow}, a sophisticated constraint reformulation tool supporting \textsc{Essence Prime} that is able to generate CSP instances for various backend solvers.

\section{Related Work}
Iori et al.~\cite{Iori} describe the \textit{Two-Dimensional Orthogonal Packing Problem} (2D-OPP) and the \textit{Two-Dimensional Knapsack Problem} (2D-KP). Both of these problems are well-studied in the field of constraint programming~\cite{Korf, deQueiroz}. A Factorio blueprint has a limited area in which to place rectangular objects on a discrete square grid. To increase the production rate, players must add more objects to a factory.
If there is no feasible packing of these objects in the blueprint, then it is impossible to build such a factory.
Checking whether 2D-OPP can be satisfied for a given set of objects can therefore be used to give an upper bound on the number of objects that can be placed in the blueprint.
An efficient factory should also maximise the number of objects which increase the production rate, such as assemblers, and minimise the use of other objects such as conveyor belts. Similarly to 2D-KP, by weighting objects by their contribution to the production rate, the most desirable set of objects can be found.

Peng et al.~\cite{Peng} proposed a solution for the \textit{Non-Crossing Multi-Agent Pathfinding} (NC-MAPF), which is the problem of finding paths for multiple agents such that every agent reaches its desired destination and their paths do not cross.
This problem closely aligns with the conveyor belt layout problem we face.
However, conveyor belts carry additional constraints. For example, paths carrying the same object may be merged together so the number of belts used is reduced.
Yu and LaValle~\cite{Yu} show that MAPF on graphs can be reduced to a network flow problem. When maximising the flow rate of products, we can trace some similarities with the railway traffic flow problem presented by Harris and Ross in 1955~\cite{HarrisRoss, FordFulkerson}.
However, we face some additional complexities, such that Factorio blueprints may have multiple types of items whose routes must be kept separate, and multiple source nodes. These differences mean existing maximum flow optimisation techniques alone are not sufficient to optimise a Factorio blueprint, but their fundamental concepts are likely to be relevant.

\section{Factorio}
\label{sec:factorio}

A brief introduction to Factorio can be found in \href{https://www.youtube.com/watch?v=KVvXv1Z6EY8&ab_channel=Factorio}{the 2016 gameplay trailer} \cite{FactorioTrailer}.
The game contains many interconnected systems that must be aligned in order to create a successful factory.
%
Factorio is played in an infinite two-dimensional grid of square \textit{tiles}.
Each tile can contain natural resources such as trees, stone, ore, and water.
These resources are the basic ingredients needed to produce all other items, so the player is encouraged to find a starting location in close proximity to as many as possible.
However these resources are finite and as the game progresses, players will exhaust what they have nearby and must expand their factory outward.
Tiles may also contain obstacles such as cliffs which cannot be traversed or built across, providing an additional challenge when expanding.
%
In addition to providing their factory with the necessary resources, players must generate electricity, defend against enemies, and more.
The game's systems are configured such that players must care about more than the production rate of their factory to play effectively.
An example is limiting the rate at which the factory pollutes the environment because pollution scales the difficulty of enemy attacks.

To make the development of a constraint model feasible, a minimal set of the game's buildings are considered from which a functional item-based blueprint can be produced:
%
\begin{figure}[tp]
     \centering
    \begin{minipage}{0.31\textwidth}
        \centering
        \includegraphics[width=\textwidth]{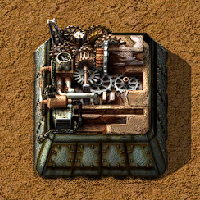}
        \caption{An assembler in-game\footnotemark[1] \footnotetext[1]{\url{https://wiki.factorio.com/Assembling_machine_1}}}
        \label{fig:assembler}
    \end{minipage}
    \hfill
    \begin{minipage}{0.31\textwidth}
        \centering
        \includegraphics[width=\textwidth]{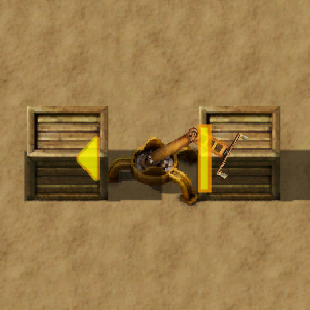}
        \caption{An inserter in-game\footnotemark[2] \footnotetext[2]{\url{https://wiki.factorio.com/Inserter}}}
        \label{fig:inserter}
    \end{minipage}
    \hfill
    \begin{minipage}{0.31\textwidth}
        \centering
        \includegraphics[width=\textwidth]{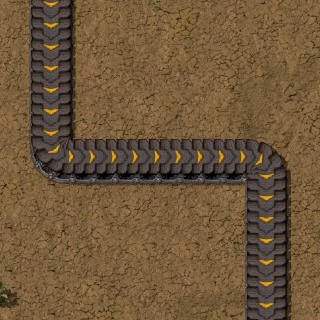}
        \caption{A path of conveyors in-game\footnotemark[3] \footnotetext[3]{\url{https://wiki.factorio.com/Transport_belt}}}
        \label{fig:conveyor}
    \end{minipage}
\end{figure}
\noindent
%
\begin{itemize}
    \item 
\textbf{Assemblers}~(\Cref{fig:assembler}) take one or more items as input and produce a new, more complex item as output.
The player must select a \textit{recipe} for the assembler to execute, which indicates the types and quantities of the items consumed and produced by the assembler.  
Each recipe produces a single type of item, and each item is the product of at most one recipe. This would not be true if fluids were considered in addition to items, which we omit for simplicity.  
Products are stored in the assembler's internal inventory and can only be removed manually by the player in real-time or automatically by an inserter. 
Ingredients can also be provided in these ways. 
We assume that there are no real time interactions between the player and the game, and that there are no limits to the assembler's inventory size. 
In the game world, an assembler is 3 by 3 tiles in size.
\item 
\textbf{Inserters}~(\Cref{fig:inserter}) are robotic arms that pick items up from their input tile and place them on their output tile, which is opposite of the input tile. The \textit{direction} of an inserter is the direction from the input to the output tile.
Inserters can access the inventories of assemblers and conveyors.
An inserter can hold one item at a time, transporting up to 50 items per minute.
If the output tile is not accepting items due to its capacity being full, the inserter will keep hold of the item and wait for the tile to start accepting items again. 
They occupy a single tile in the game world. 
\item 
\textbf{Conveyors}~(\Cref{fig:conveyor}) transports items placed on them from one tile to another.
They can be connected together to create long routes.
In game, conveyors carry items in two lanes. However, for simplicity, we shall assume that there is only one lane. 
At maximum capacity, a conveyor lane can transport 450 items per minute.
This is a much higher throughput than an inserter, allowing multiple inserters to interact with a single line of conveyors.
\end{itemize}

\paragraph*{The blueprint problem}

A \textit{blueprint} is a layout of a rectangular area that can be copied. 
Blueprints typically contain unterminated transport networks that the player is expected to connect to the rest of their factory in order to provide the needed resources.
Since these copies can be connected to form bigger factories, there are indications of the tiles where the items go into and out of the blueprint area (which we shall refer to as \textit{blueprint sources} and \textit{blueprint destinations} respectively), as well as their types.
For simplicity, we will consider blueprints with possibly multiple sources but only a single destination. 
The problem we are trying to model is one where the dimension of this blueprint area and the ingoing-outgoing items types and positions are given beforehand. 
Furthermore, the input rate of the blueprint sources (in the number of items per minute) is also given. 

The blueprint problem aims to find a layout of assemblers, inserters and conveyors which takes the input items and produce the required output items in the specified places. 
The recipe for each assembler used should also be determined, which are chosen from a library of available recipes.
Needless to say, the layout must respect the game mechanics. 
In addition to that, we are also interested in finding the optimal layout. We define optimality in this context as a layout which maximises the production rate (number of outgoing items in a minute) and minimises the cost. The cost is related to the number of buildings needed. We shall give further detail to our objective functions later.


\section{The Model}
\label{sec:model}

This section starts by describing the model parameters and constants in \Cref{main model params}. In \Cref{gameRep} we describe the viewpoint taken for modelling the core components of Factorio. Given preliminary experimental results, a single optimisation model was shown to be infeasible for solving non-trivial blueprints. 
Therefore, in the spirit of a Benders decomposition approach, we implemented a multi-stage solution that splits the problem into three models, which is described in \Cref{subsec:componentModels}. 

\subsection{Model Parameters and Constants}
\label{main model params}

The width $w$ and height $h$ of the blueprint are given to the model as two integers \lstinline|binW| and \lstinline|binH|. 
In all models, we define \lstinline|xs| and \lstinline|ys| to be the domains \lstinline|int(1..binW)| and \lstinline|int(1..binH)| respectively.  

The number of item types present in the blueprint is given using an integer parameter \lstinline|num_items|.
The items types are identified as integers from 1 to \lstinline|num_items|. 
The integer parameter \lstinline|out_item| indicates the type of item the blueprint should output. 

The blueprint destination is given in a $0/1$ matrix \lstinline|outputs| of size $w \times h$.
The blueprint sources are given in a matrix \lstinline|input_items| of the same size, whose entry indicate the item type (as an integer) that is available in the corresponding tile.
Similarly, each entry of \lstinline|input_qtys| indicates the number of items of the specified type that is available in the corresponding tile in a minute. 

The recipes are described in \lstinline|recipe_qtys|, a matrix of size \lstinline{input_items} $\times$ \lstinline{input_items}. For $i \neq j$, the $(i,j)$-th entry in the matrix indicates the number of Item $j$ needed to produce Item~$i$. The number of Item~$i$ produced is indicated in the $(i,i)$-th entry of \lstinline|recipe_qtys|, and the rate of production (in number of items produced per minute) is indicated in the $i$-th entry of the parameter \lstinline|recipe_rates|, a list of length \lstinline{input_items}. 
To give the user extra control over the model, the user can also indicate the rate of item transfer of an inserter, in an integer parameter \lstinline|inserter_rate|. 





\subsection{Representing the Game World}
\label{gameRep}






\paragraph*{Modelling conveyors and inserters routes}



\begin{figure}[tp]
    \centering
    \begin{minipage}[t]{0.3\textwidth}
        \centering
        \includegraphics[width=\textwidth]{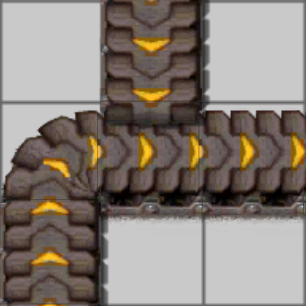}
        \caption[Example transport route to demonstrate possible representations]{In game representation of a conveyor route. }
        \label{fig:conveyor_rep}
    \end{minipage}
    \hfill
    \begin{minipage}[t]{0.3\textwidth}
        \centering
        \begin{tikzpicture}[scale=\textwidth/1cm]
            \draw[step=0.333cm,color=black] (0,0) grid (1,1);
            \node at (0.166,0.833) {.};
            \node at (0.500,0.833) {1};
            \node at (0.833,0.833) {.};
            \node at (0.166,0.500) {2};
            \node at (0.500,0.500) {3};
            \node at (0.833,0.500) {4};
            \node at (0.166,0.166) {1};
            \node at (0.500,0.166) {.};
            \node at (0.833,0.166) {.};            
        \end{tikzpicture}
        \caption[Incremental conveyor representation]{Incremental representation of \Cref{fig:conveyor_rep}.}
        \label{fig:conveyor_rep_inc}
    \end{minipage}
    \hfill
    \begin{minipage}[t]{0.3\textwidth}
        \centering
        \begin{tikzpicture}[scale=\textwidth/1cm]
            \draw[step=0.333cm,color=black] (0,0) grid (1,1);
            \node at (0.166,0.833) {.};
            \node at (0.500,0.833) {S};
            \node at (0.833,0.833) {.};
            \node at (0.166,0.500) {E};
            \node at (0.500,0.500) {E};
            \node at (0.833,0.500) {E};
            \node at (0.166,0.166) {N};
            \node at (0.500,0.166) {.};
            \node at (0.833,0.166) {.};            
        \end{tikzpicture}
        \caption[Directional conveyor representation]{Directional representation of \Cref{fig:conveyor_rep}.}
        \label{fig:conveyor_rep_dir}
    \end{minipage}
\end{figure}

Two representations of the conveyors are considered: the \textit{incremental representation}, which labels conveyors incrementally as the route advances towards the destination; and the \textit{directional representation}, which labels conveyors based on the direction of flow (where integers 1 to 4 are used to indicate North, South, East and West respectively). See \Cref{fig:conveyor_rep,fig:conveyor_rep_dir,fig:conveyor_rep_inc} for an example, where the top-centre and bottom-left tiles are sources, and the right-middle tile is a destination. Like conveyors, inserters must face in one of four directions, meaning these representation can also be used. 

More specifically, the incremental representation, similarly described in~\cite{GebserJR14}, comes with the following constraints:
\begin{enumerate}
    \item Source tiles contain a 1;
    \item Destination tiles must be greater than 1;
    \item The tile with the largest value must be a destination tile;
    \item A non-zero value in a tile that is not a destination implies a larger value exists in an orthogonal tile.
\end{enumerate}
As we shall see later, we will be attempting to form routes to connect buildings in the blueprint. So the source and destination here may not correspond to the blueprint source and blueprint destination defined in \Cref{sec:factorio}. 
A critical challenge to overcome is the prevention of cycles, where sources or destinations are self-connected. 
The fourth constraint of the incremental representation inherently prevents cycles forming because the end of the cycle will have a larger value than the start. 
It also allows multiple paths to merge, since the constraint only specifies a `larger value' instead of a `consecutive value'.
However, this representation does not scale efficiently. 
Our solution to the issue is to combine the  incremental and directional representations to allow efficient cycle breaking and easy access to the directions. 
The variable \lstinline{routes} represents the combination of conveyors and inserters that brings items from given sources to given destinations. 
\Cref{routes} gives the viewpoint and the channelling constraint used.
\begin{lstlisting}[caption={Combined incremental and directional representation for transport routes.},label=routes]
find conveyors : matrix indexed by [ys, xs] of int(0..4)
find inserters : matrix indexed by [ys, xs] of int(0..4)
find routes    : matrix indexed by [ys, xs] of int(0..binW+binH)
forall y : ys . forall x : xs .  
    routes[y, x] > 0 <-> (conveyors[y, x] > 0 \/ inserters[y, x] > 0)
\end{lstlisting}

We require all conveyors and inserters to belong to a transport route. A route either starts from a conveyor on a  blueprint source tile or from an inserter taking items out of an assembler. So these starting tiles will have \lstinline|1| in the corresponding entries in \lstinline|routes|. Conveyors or inserters which do not start a route increment the longest route they are part of. 
Inserters can only receive input from one space, but conveyors can receive input from multiple routes.
By ensuring all directions providing input have a smaller route value than the current space, the longest route is extended, producing the expected behaviour.


Note that the domain of \lstinline{routes} is limited by $w+h$ instead of the area $w \times h$. 
A naive upper bound of the domain $w \times h$ does not scale well. 
While it is true that you can get from a source to any destination in $w+h$ steps, it may not be true with the presence of obstacles. 
Since the only possible obstacles are assemblers and inserters, we conjecture that a solution which requires routes longer than $w+h$ can be rearranged so that the routes are within the $w+h$ bound. 
Furthermore, a solution that maximises the output rate and minimises the cost would likely need an efficient routing system, limiting the scope for long routes.  
Inserters have a lower item throughput than conveyors, so to maximise item flow in the factory, they should not be used when a conveyor would suffice. 
Therefore, our objective function punishes overuse of inserters (see \Cref{objectiveForRoutes}). The penalties can take any value as long as we have \lstinline|conveyor_penalty| $>$ \lstinline|inserter_penalty|. 
\begin{lstlisting}[caption={Objective function for routing to minimise the number of inserters.}, label={objectiveForRoutes}]
minimising sum y:ys. sum x:xs. (conveyor_penalty * (conveyors[y, x ] > 0)) 
                               + (inserter_penalty * (inserters[y, x] > 0))                
\end{lstlisting}

\paragraph*{Modelling item flow}


To determine if a blueprint is valid, the model must track which items are present on which tiles.
Constraints can then be placed on inserters passing ingredients to assemblers to ensure that only items required by the assembler's recipe are being provided.
Similarly, assemblers can be constrained to ensure all ingredients required by their recipe are provided.

In Factorio, multiple types of item can be present on the same conveyor.
Sometimes the quantity of an item is not sufficient to use the entire throughput of the conveyor, and if multiple low-quantity items are being transported to the same place, it can be more space efficient to place them all on the same conveyor.
This increases the density of conveyor networks, potentially allowing more assemblers and inserters to be packed into the space and improving the production rate of the factory.
The cost of properly modelling this feature is likely to be too expensive.
Allowing any and all items to be present on each tile requires a 3D matrix of size \lstinline|num_items| $\times$ \lstinline|binW| $\times$ \lstinline|binH|, or an explicit equivalent, which scales poorly.

A compromise is made to ensure the model is of an acceptable complexity: each tile may only carry one type of item.
This causes optimality to be lost as the conveyor-density improvement discussed above is no longer available.
Fortunately it is unlikely that this increased density will provide enough additional, usable space to significantly increase the performance of a factory except in very specific edge cases. 
It is rare for more than two items to be present on a given conveyor due to inserters picking up any item they come across, and efficient factories will typically devote entire transport routes to the most commonly used items due to their high demand.

\begin{figure}[]
    \centering
    \begin{subfigure}[t]{0.42\textwidth}
        \centering
        \includegraphics[width=\textwidth]{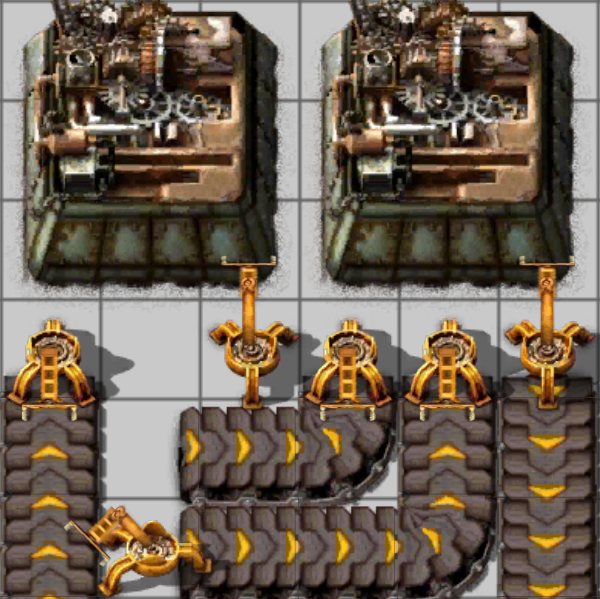}
        \caption{The in-game illustration of the blueprint. }
        \label{fig:carrying}
    \end{subfigure}
    \hfill
    \begin{subfigure}[t]{0.42\textwidth}
        \centering
        \begin{tikzpicture}[scale=\textwidth/1cm]
            \draw[step=0.1667cm,color=black] (0,0) grid (1,1);
            \node at (0.083,0.916) {2};
            \node at (0.250,0.916) {2};
            \node at (0.416,0.916) {2};
            \node at (0.583,0.916) {3};
            \node at (0.750,0.916) {3};
            \node at (0.916,0.916) {3};

            \node at (0.083,0.750) {2};
            \node at (0.250,0.750) {2};
            \node at (0.416,0.750) {2};
            \node at (0.583,0.750) {3};
            \node at (0.750,0.750) {3};
            \node at (0.916,0.750) {3};
            
            \node at (0.083,0.583) {2};
            \node at (0.250,0.583) {2};
            \node at (0.416,0.583) {2};
            \node at (0.583,0.583) {3};
            \node at (0.750,0.583) {3};
            \node at (0.916,0.583) {3};

            \node at (0.083,0.416) {1};
            \node at (0.250,0.416) {.};
            \node at (0.416,0.416) {2};
            \node at (0.583,0.416) {2};
            \node at (0.750,0.416) {1};
            \node at (0.916,0.416) {3};
            
            \node at (0.083,0.250) {1};
            \node at (0.250,0.250) {.};
            \node at (0.416,0.250) {2};
            \node at (0.583,0.250) {2};
            \node at (0.750,0.250) {1};
            \node at (0.916,0.250) {3};

            \node at (0.083,0.083) {1};
            \node at (0.250,0.083) {1};
            \node at (0.416,0.083) {1};
            \node at (0.583,0.083) {1};
            \node at (0.750,0.083) {1};
            \node at (0.916,0.083) {3};
        \end{tikzpicture}
        \caption{An illustration of the decision variable \lstinline|carrying|, which indicates which item is present on each tile.}
        \label{fig:carrying_rep}
    \end{subfigure}
    \caption[Blueprint to demonstrate item carrying]{shows a blueprint containing two assemblers.
The assembler on the left takes Item~1, provided to the blueprint via the bottom-left tile, and produces Item~2.
The assembler on the right takes Items~1 and 2 and produces Item~3, which is outputted from the blueprint from the bottom-right tile.}
\label{fig:itemflow}
\end{figure}

We model item flow as a matrix \lstinline|carrying| indexed by \lstinline|[ys, xs]| of \lstinline|int(0..num_items)|, with the following constraints (see \Cref{fig:itemflow} for an example). 
\begin{enumerate}
    \item Tiles carry an item if and only if there is a game object present.
    \item Conveyors carry all items passed to them from adjacent tiles.
    \item Inserters carry any item carried on their input tile.
    \item Assemblers carry the item they produce.
\end{enumerate}


\paragraph*{Modelling item flow rate}

One of our aims is to maximise the production rate. This requires knowing the item flow rate through each conveyor tile. 
However, recipe duration and item flow rates are both continuous variables. 
Approximations and unit conversions can be used to convert continuous variables to discrete variables at the cost of resolution. 
The representation used should be able to model an item with a recipe duration of 1 item per minute, but also a full conveyor lane transporting 7.5 items per second, or 450 items per minute.
In our experiments, a domain of \lstinline|int(1..450)| is too large to be applied over every tile of the blueprint.
There is no way of modelling continuous item rates accurately in \textsc{Essence Prime} without causing state explosion.
Any representation which does not cause state explosion achieves this by reducing the resolution to a point of unusability. A unified constraint optimisation model (COP) model was therefore deemed not feasible. 
Hence we decided to proceed with a multi-stage COP model.

\subsection{Multi-Stage COP Model}
\label{subsec:workflow}
\label{subsec:componentModels}



A multi-stage model allows the modelling of item rates and the blueprint's space to be handled separately.
If the routing stage can be given the optimal layout of assemblers and inserters as a parameter, the problem becomes one of pathfinding again, and not of flow rates.
In fact, it appears very similar to the problem investigated by Reid et al. \cite{Reid}, but involving multiple paths.
For each transport route, all other routes provide the obstacles that must be routed around.
The model attempts to find an assignment which allows all routes to reach their destination, if such an assignment exists.

\begin{figure}[t]
    \centering
    \includegraphics[height=0.45\textheight]{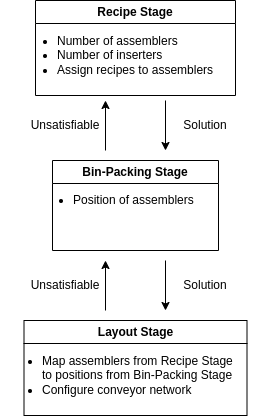}
    \caption{The structure of the multi-stage model.}
    \label{fig:stage_flowchart}
\end{figure}

This must be the final stage, since the layout of assemblers and inserters must be known.
A stage deciding this layout also needs to model the 2D space of the blueprint, so it cannot also model item rates.
It could model a bin-packing problem in which the number of assemblers and inserters are given, and a valid packing in the blueprint must be found.
This would be an extension of classic packing problems because inserters need to be adjacent to assemblers. 
This leaves an initial stage, which decides the number of assemblers and inserters, and the recipes used.
The model does not need to care about where the objects are placed in the blueprint, as long as the area they occupy does not exceed its area.
Figure~\ref{fig:stage_flowchart} illustrates the purpose of each stage and how they interact with each other.
In summary, we have:
\begin{enumerate}
    \item \textbf{Recipes Stage}: This model takes recipes and items as parameters, and
finds the number of assemblers and the recipe followed by each which maximises the production rate.
Additionally, knowing how much each item needs to be provided to each assembler to achieve this optimal rate, the number of inserters attached to each assembler is found.
    \item \textbf{Bin-packing Stage}: This stage uses the number of assemblers and the number of inserters attached to each from the Recipe Stage, to ensure it can fit in the blueprint. An important note is that this stage does not take into account the item types. Meaning that we do not take into account which recipes are assigned to which assemblers, and what items the inserters carry.  
  \item \textbf{Layout Stage}: This stage takes the assembler positions and the number of inserters attached to each assembler to generate a routing system so that all assemblers receive the items they need. This means that the output of Stage 2 does not fix the locations of the inserters interacting with the assemblers. Furthermore, the assemblers in Stage 2 have not correlated with the assemblers in Stage 1, so we shall do that in this stage also.
\end{enumerate}

The parameters for each stage can be generated programmatically based on the output of the previous stages. 
In the case of the first stage, its parameter file is generated from the parameters described in \Cref{main model params}, which we shall refer to as the \textit{main parameters}.
If a stage is unsatisfiable, the previous stage must generate a new solution, continuing until the final stage is satisfied or all solutions have been attempted.

The models, the code to construct the parameters and some sample inputs and outputs can all be found in the GitHub repository \cite{factorioRepo}.

\subsubsection{Stage 1: Recipes}


The stage parameters are taken from the main parameters using little to no preprocessing.  
Two additional parameters are calculated from the main parameters to give finite domains for decision variables: 
\lstinline|max_rate| is calculated as the maximum of the recipe rates; \lstinline|max_assemblers| is calculated as $ \lfloor w / 3\rfloor \times  \lfloor h / 3\rfloor$, which gives us the finite domain 
\lstinline|assemblers|, defined to be \lstinline|domain int(1..max_assemblers)|.

The model aims to find the number of assemblers needed, the recipes assigned to each assembler and their productions rates. 
These forms the decision variables \lstinline|num_assemblers|, \lstinline|assembler_recipes| and \lstinline|assembler_rates|. 
The recipes are identified by the item it produces, since there are at most one recipe that produce a given item. 

The number of assemblers and the maximum possible assemblers may be different.
Assemblers which could theoretically be present but are not used in the solution are called \textit{inactive} assemblers.
They are given a recipe 0, which does not correspond to an item.
Some symmetry is broken by forcing inactive assemblers to the end of the array.

The model should also decide the number of inserters carrying items in and out of the assemblers, and the item type they carry.  
Intuitively, the number of inserters matches the ratio of items in the recipe, taking into account the  rate at which each item is present.
The ingoing inserters are tracked in a decision matrix \lstinline|inserters_in| indexed by \lstinline|[assemblers, items]|. Since each assembler has only one output item type, the outgoing inserters are tracked with an array \lstinline|inserters_out| of indexed by \lstinline|assemblers|.

Each assembler can be assigned to up to 12 inserters (3 on each side), which suggests a domain of of size 12. However, \lstinline|max_inserters_in| is calculated from the recipes to restrict the domain (see \Cref{maxconsumption}). 
The main idea is that if no item is consumed in a quantity greater than $k$ inserters can provide, there is never a need for more than $k$ inserters providing that item to an assembler. The domain of \lstinline|inserters_out| is bounded by \lstinline{max_inserters_out}, which is calculated similarly. 

\begin{lstlisting}[caption={Limiting the number of inserters transporting items into an assembler. }, label={maxconsumption}]
letting max_consumption = max(
   [ (recipe_qtys[i, j]*recipe_rates[i]) |  i: items, j: items, i != j ] )
letting max_inserters_in  =  (max_consumption / inserter_rate)
                              + ((max_consumption % inserter_rate) != 0) 
\end{lstlisting}

Knowing the consumption rate (how much of each item is demanded by each assembler) is important for calculating production rates.
This is kept track of using the decision variable \lstinline|consuming|, a matrix indexed by \lstinline|[assemblers, items]| of \lstinline|int(0..max_consumption)|. 


If the solution to this stage does not have a valid packing in the next stage, this stage will be repeated to find the next solution.
To prevent the same solution being generated each time, parameters describing the previous solutions are given so constraints can be written to guarantee uniqueness (see \Cref{storeAttemptsStage1}).
\begin{lstlisting}[caption={Parameters describing solutions that have already been considered in the later stages.}, label={storeAttemptsStage1} ]
given num_attempts : int(1..)
given assembler_attempts :
  matrix indexed by [int(1..num_attempts), assemblers] of int(0..num_items) 
given inserter_attempts :
  matrix indexed by [int(1..num_attempts), assemblers] of int(0..12)
\end{lstlisting}
The inserter counts are given as an overall sum for each assembler because changing which item each inserter interacts with will not cause a valid packing to exist where one previously did not.
The assignments of \lstinline|assembler_recipes|, \lstinline|inserters_in| and \lstinline|inserters_out| are checked against the previous solutions.
If it is the same as any of them, it is rejected.

The objective function aims to maximise the production of the output item.
A small penalty is applied so that solutions achieving the same rate with fewer assemblers and inserters are prioritised (see \Cref{stage1objectiveFunction}). 
\begin{lstlisting}[caption={The objective function maximises the production rate and penalises number of buildings.}, label={stage1objectiveFunction}]
maximising (sum i: assemblers. 
                  (assembler_rates[i] * (assembler_recipes[i] = output) )   
           - (sum i: assemblers. (9 * (assembler_recipes[i] > 0))
                                 + (sum item: items. inserters_in[i, item])
                                 + inserters_out[i]    )         
\end{lstlisting}


Most of the game rules can then be modelled as constraints in a straightforward way. Some notable constraints regarding the production and consumptions rules and symmetry breaking are described below: 
\begin{itemize}

\item  The consumption rate of an item cannot exceed its production rate, because the item must exist in the blueprint in order to be consumed. Likewise, the production rate of an item cannot exceed its recipe rate.

\item  If a number of inserters are assigned to interact with an assembler, all of the inserters should be handling items at a non-zero rate. Furthermore, all inserters should operate at their maximum rate, excluding one inserter handling the remaining items in case their flow rate is not a multiple of the inserter rate.
This means that for an assembler with $n > 0$ inserters and inserter rate $r$, the rate at which the assembler's production rate must be between $nr$ and $(n - 1)r$. Similar is true for the consumption rate. 


\item  An ordering is forced on the decision variables to eliminate symmetry.
Active assemblers are ordered by the item they produce, with ties broken by their inserter count.
Each solution now represents a unique assignment of assemblers and inserters.

\end{itemize}

\subsubsection{Stage 2: Bin-Packing}


This stage is given the number of assemblers (\lstinline|num_assemblers|),  along with their number of attached inserters (as an array \lstinline|inserters| of length  \lstinline|num_assemblers|).
The blueprint sources and destination tiles are given as an 0/1 matrix \lstinline|reserved| indexed by \lstinline|[ys, xs]|. 


The decision variable \lstinline|assembler_layout| indicates the tiles containing assemblers using the occurrence representation~\cite{occurrence}: an 0/1 matrix the same dimension as the blueprint. 
The presence of an assembler is indicated with a $1$ in the top-left tile containing the assembler. 
To disambiguate which assembler an inserter is attached to, the axis of movement is needed, but the exact direction is not important. 
This can be modelled as \lstinline|horizontal = 1| and \lstinline|vertical = 2|.
So the decision variable \lstinline|inserter_layout| representing the inserter layout is similar to \lstinline|assembler_layout|, but with \lstinline|int(0..2)| as its domain. 
Additionally, an auxiliary decision variable \lstinline|positions|, which maps the assemblers to their attached inserters, is included for more succinct constraints. 

Like in the previous stage, it is possible that solutions of this stage are rejected by the next stage.
The model must remember previous solutions to avoid recreating them.
Only the assembler layout is stored because the next stage will check all possible arrangements of inserters around each assembler (see \Cref{storePrevSolnsStage2}). The \lstinline|attempts| matrix represents each layout as a flattened array of length $wh$. Each coordinate $(x, y)$ is mapped to index $(y - 1)w + x$.

\begin{lstlisting}[caption={Making sure that the solution is different from previous attempts.}, label={storePrevSolnsStage2}]
given num_attempts : int(1..)
given attempts : matrix indexed by 
                  [int(1..num_attempts), int(1..binW * binH)] of int(0..1)
forall n : int(1..num_attempts) .
    flatten(assembler_layout) <lex attempts[n, ..]   \/ 
    flatten(assembler_layout) >lex attempts[n, ..],
\end{lstlisting}



Other constraints are straightforward and available from the linked repository. 

\subsubsection{Stage 3: Layout}


    
    


The parameters are taken directly from the solutions of the previous stages.
They describe the blueprint dimension, the blueprint source and destination locations and item types, the total number of item types, the recipes, the number of assemblers, and 
for each assembler: which recipe it follows; how many inserters provide each item type; and how many inserters retrieve the product from the assembler.
%
This stage aims to generate a transport route so that all assemblers receive the items they need. 
The decision variables regarding routing and item carrying are as described in \Cref{gameRep}. 
Note that Stages 1 and 2 only model inserters which interact with assemblers.
Inserters used for splitting transport routes were not considered. 

The \lstinline|assembler_layout| parameter, taken from the solution of Stage 2, defines which tiles contain assemblers, but not which assemblers they contain.
This increases the versatility of each solution from the previous stage since all possible assignments of assemblers to positions are evaluated, reducing the amount of communication between the stages.
A decision variable \lstinline|assignments| is declared to track the mapping between the assemblers and \lstinline|assembler_layout|.
Finally, as noted above, the positions of the inserters interacting with the assemblers are not fixed in Stage 2, we merely check that they can fit in the blueprint area. In this stage, we shall 
determine the final positions of these inserters.

The constraints modelling the routes and item flow are described in \Cref{gameRep}. Other constraints are straightforward. 
Because item flow rates are no longer modelled, there is a risk that optimality is lost in this stage.
The model does not know how much of each item is travelling along each route, so if a route is split, the items may not be distributed by inserters in the optimal ratio.
It is assumed that the optimal layout is so densely packed that there is limited scope for this to occur.

\section{Result and Discussion}
\label{sec:result}

We have described a working solution for a new and interesting problem. 
The models, the code to construct the parameters and some sample inputs and outputs can all be found in the GitHub repository \cite{factorioRepo}. 
The presented solution is able to find solutions for non-trivial problems, and could be effectively used as a game companion.
Preliminary experiments showed that a unified model able to tackle non-trivial inputs is not easily achievable. We identified the need to represent the flow rates in \textsc{Essence Prime} as the main reason why our unified model approach has not worked.

The developed multi-stage approach needs to generate parameter files, read solutions, and step back to previous stages when an attempt fails.
This back-and-forth communication between each stage is detrimental to the model's performance, as there are limits to what information can be shared between stages. This weakness identifies a potential area for future work.


More concretely, blueprints with an area of around 100 tiles can be generated in several minutes.  
Beyond this size, runtimes increase rapidly and soon become intractable due to the number of failed attempts and subsequent retrying of each stage.
Given that there are hundreds of items in Factorio that the player needs to produce to complete the game, it is clearly unfeasible to optimise a complete factory.
However, 100 tiles is enough to contain several assemblers, meaning if each intermediate item is given its own blueprint, a modular factory could be designed using this model.
Clearly, the factory generated by this approach is not necessarily optimal. Still, we conjecture that each sub-factory within it being optimal would mean that the overall result would not be far from it. 

Solutions generated by the presented approach have been closely inspected and observed to be correct in the sense that they can be built in Factorio and will generate the intended product, because the simplifications made do not introduce new behaviours to the model that are not present in the game.

Removing some game features means that the rates of production predicted by the model may not be the same as the in-game production rate.
The most obvious display of this inaccuracy is when item flow rates are low.
The model cannot identify cases where inserters will be starved of input, and may consider a layout valid despite some assemblers never receiving their ingredients. This was a necessary compromise in order to develop a tractable model. 
There also exist scenarios where item distribution on conveyors is sub-optimal due to the final stage of the model not considering flow rates.
This compromise had to be made to ensure the model could solve even trivial instances.

We finally highlight that in gameplay terms, the usage of the term `optimal' is subjective.
Through the paper we have tried to optimise the throughput of the factory. Still, some players might instead prefer blueprints that can be scaled easily at the cost of maximum efficiency. 

\subsection{Case Studies}

\begin{figure}[tp]
    \centering
    \begin{subfigure}[t]{0.3\textwidth}
        \centering
        \includegraphics[width=\textwidth]{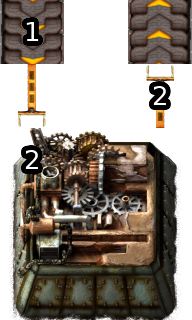}
        \caption {Recipe requires Item 1 and produce Item 2 in 1:1 ratio.}
        \label{fig:3x5_solution1-1}
    \end{subfigure}
    \hfill
    \begin{subfigure}[t]{0.3\textwidth}
        \centering
        \includegraphics[width=\textwidth]{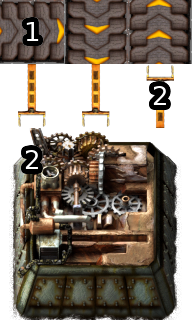}
        \caption {Recipe requires Item 1 and produce Item 2 in 2:1 ratio.}
        \label{fig:3x5_solution2-1}
    \end{subfigure}
    \hfill
    \begin{subfigure}[t]{0.3\textwidth}
        \centering
        \includegraphics[width=\textwidth]{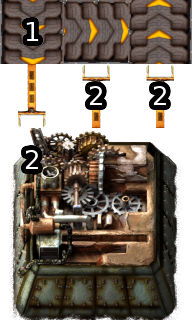}
        \caption {Recipe requires Item 1 and produce Item 2 in 1:2 ratio.}
        \label{fig:3x5_solution1-2}
    \end{subfigure}
    \caption{shows that the ratio of the ingoing and outgoing inserters adapt to different recipes.}
    \label{fig:3x5_solution}
\end{figure}

We now illustrate the success of our approach with some example outputs. 
\Cref{fig:3x5_solution1-1,fig:3x5_solution2-1,fig:3x5_solution1-2} show outputs for a minimal 3x5 factory with assembler recipes having, respectively,  1:1, 2:1 and 1:2 ratios between the ingredient (Item 1) and the product (Item 2).
The number of inserters carrying items into and out of the assembler matches the ingredient-product ratio of the recipe so that the assembler can work at full capacity, while also not having more ingoing items than it could process nor more outgoing items than its production rate. 

\begin{figure}[tp]
    \centering
    \begin{minipage}{0.32\textwidth}
        \centering
        \includegraphics[width=\textwidth]{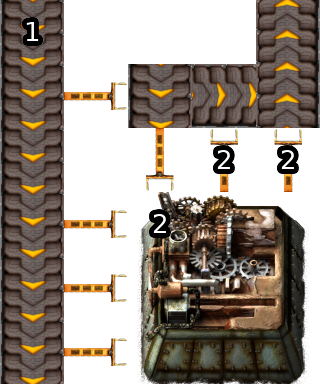}
        \caption{Result for a 5x6 blueprint, where the top-left and the top-right tiles are the blueprint source and destination tiles respectively. }
        \label{fig:5x6_solution}
    \end{minipage}
    \hfill\vline\hfill
    \begin{minipage}{0.66\textwidth}
        \centering
        \begin{subfigure}[t]{0.49\textwidth}
            \centering
             \includegraphics[width=\textwidth]{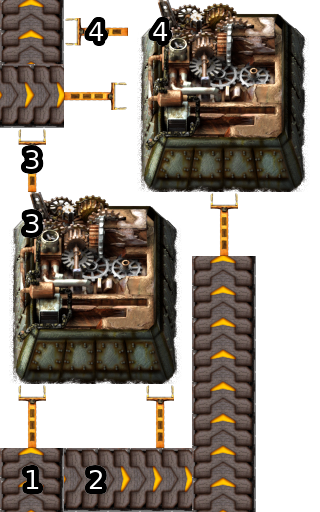}
            \caption{Item 2 is provided in the tile to the right of the bottom-left tile. }
            \label{5x8blockingInput}
        \end{subfigure}
        \hfill
        \begin{subfigure}[t]{0.49\textwidth}
            \centering
            \includegraphics[width=\textwidth]{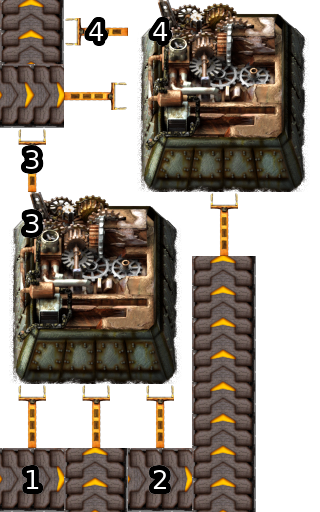}
            \caption{Item 2 is provided in the tile two tiles to the right of the bottom-left tile. }
            \label{5x8UnblockedInput}
        \end{subfigure}
        \caption{Results for 5x8 blueprint instances where Item 1 is provided in the bottom-left corner, but with different Item 2 source locations. }
        \label{fig:5x8_solution}
    \end{minipage}
\end{figure}

Figure \ref{fig:5x6_solution} shows an output for a 5x6 instance with a 2:1 recipe ratio.
As in the previous example, the number of inserters matches the ratio of the recipe. 
This blueprint is large enough to contain two assemblers, but it is not possible to configure a valid factory with two assemblers, given the input and output locations.
This example shows that the model is able to handle situations where the optimal number of assemblers differs from the maximum.


Figure \ref{fig:5x8_solution} exhibits several interesting behaviours.
The output of one assembler is used by another, showing that intermediate products can be created and transported for use as an ingredient.
Item 2 is demanded by two assemblers executing different recipes, which are served by a continuous conveyor route, exhibiting the branching of transport routes. 
Note that if Item 2 is provided at a rate of \lstinline|inserter_rate| or lower, no items will reach the second inserter.
Because the layout stage does not model flow rates, this situation cannot be handled optimally by the model; the optimal outcome would be the model identifying that no valid blueprint exists.
It should also be noted that the Factorio blueprint system does not account for this - the player using the blueprint is responsible for providing items at a sufficient rate to prevent starvation.

The bottom assembler uses a recipe which takes Items 1 and 2 and produces Item 3 in a 2:1:1 ratio. 
In \Cref{5x8blockingInput}, due to the location of the source tile of Item 2, there are not enough spaces for the ingoing inserters to match the recipe ratio. However, the ratio is matched in \Cref{5x8UnblockedInput}, where the source tile for Item 2 is one tile further to the right. 


\begin{figure}[tp]
    \centering
    \includegraphics[width=0.6\textwidth]{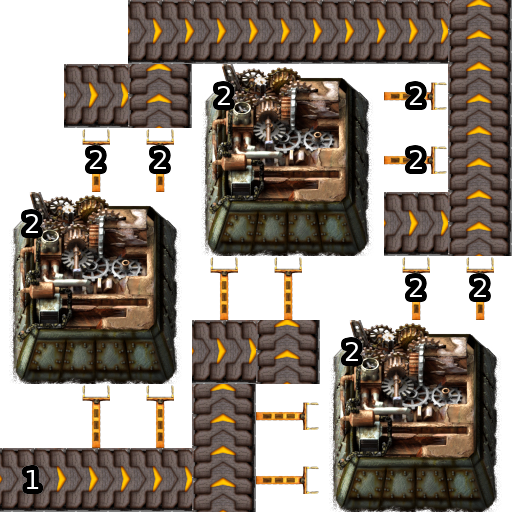}
    \caption{An unintuitive result for an 8x8 blueprint instance. }
    \label{fig:8x8_solution}
\end{figure}

Figure \ref{fig:8x8_solution} identifies complicated patterns that a human player is likely to miss.
The optimal blueprint may contain an unintuitive layout of objects in order to maximise the number of assemblers.
The objects seem randomly placed throughout the space in a pattern a human player would probably never consider, opting instead for neat rows of assemblers and straight transport routes.

Again, this raises an interesting point regarding the definition of an optimal factory.
This design is optimal for an 8x8 blueprint, but the optimal 9x9 blueprint for the same items and recipes could look very different.
Given that Factorio focuses heavily on expansion, an easy-to-expand but slightly suboptimal blueprint might be preferred to an optimal blueprint that cannot be expanded.
The concept of a blueprint being tileable is difficult to accurately define and therefore constraint. 

\subsection{Future Work}
In all examples analysed, most of the blueprint space is occupied by buildings.
This matches the assumption made previously that optimal blueprints are likely to be densely packed, limiting the available space for transport routes longer than $w + h$ tiles. A unified model would allow us to mechanically search for the existence of optimal layouts with routes longer than $w+ h$ tiles. 

A solving technique that supports modelling continuous variables would greatly simplify the model, hopefully allowing it to operate in a single stage.
SMT solvers such as Z3\cite{Z3} have the expressive power to do so, and therefore it is something worth exploring. 

More features of the game could be added to the model.
The implementation of electrical grids, underground conveyors, and more would enrich the model, making its output more applicable for direct use in Factorio.
However, this is also a tradeoff between fidelity and model performance.

Different metrics could also be optimised.
Instead of maximising the production rate in a given area, the model could be given a target production rate and find the smallest factory that achieves it.
This would require a different set of models, since new parameters and a different objective function would be needed.

As previously noted, an undesirable feature of the multi-stage approach is the lack of communication between models.
Therefore there are patterns identified by the solver on one stage that are not carried to the next, which can result in some fruitless search. 
Finally, the models can be improved by considering implied and symmetry breaking constraints. How these can be as effective as possible considering the multi-stage approach should be studied. 


\bibliography{references}

\end{document}